\documentclass{article}

\usepackage[english]{babel}
\usepackage[utf8]{inputenc}

\usepackage{times} 

\usepackage[preprint]{neurips_2024}

\usepackage{float} 
\usepackage[utf8]{inputenc} 
\usepackage{hyperref}       
\usepackage{url}            
\usepackage{booktabs}       
\usepackage{amsfonts}       
\usepackage{nicefrac}       
\usepackage{microtype}      
\usepackage[table]{xcolor}  
\usepackage{amsmath}        
\usepackage{enumitem}       
\usepackage{graphicx}       
\usepackage{multirow}       
\usepackage{listings}       
\usepackage{tcolorbox}      
\newcommand{\startrow}[1]{\rowcolor{white!10} \texttt{<|im\_start|>} #1  \\}

\newcommand{\backgroundrow}[1]{\rowcolor{blue!10} \texttt{<background>} #1 \texttt{</background>} \\}
\newcommand{\questionrow}[1]{\rowcolor{yellow!10} \texttt{<question>} #1 \texttt{</question>} \\}
\newcommand{\thinkrow}[1]{\rowcolor{blue!10} \texttt{<think>} #1 \texttt{</think>} \\}
\newcommand{\searchrow}[1]{\rowcolor{green!10} \texttt{<search>} #1 \texttt{</search>} \\}
\newcommand{\resultrow}[1]{\rowcolor{gray!10} \texttt{<result>} #1 \texttt{</result>} \\}
\newcommand{\shortanswerrow}[1]{\rowcolor{yellow!10} \texttt{<short\_answer>} #1 \texttt{</short\_answer>} \\}
\newcommand{\longanswerrow}[1]{\rowcolor{yellow!10} \texttt{<long\_answer>} #1 \texttt{</long\_answer>} \\}
\newcommand{\row}[1]{\rowcolor{white!10} #1 \\}
\newcommand{\rowtest}[1]{\rowcolor{yellow!10} {\centering#1}  \\}

\definecolor{mygray}{rgb}{0.94,0.95,0.95}
\lstdefinelanguage{prompt}{
    basicstyle=\normalfont\fontfamily{pcr}\selectfont,
    showstringspaces=false,
    breaklines=True,
    backgroundcolor=\color{mygray},
}
\tcbuselibrary{listings,skins,breakable}
\tcbset{
    enhanced,
    breakable,
    colback=mygray, %
    colframe=black, %
    boxrule=0.2mm, %
    arc=1mm, %
    top=1pt,
    bottom=1pt,
    left=2pt,
    right=2pt,
    listing only,
    listing options={
        basicstyle=\ttfamily\small, %
        language=Java, %
    },
}

\hypersetup{
    colorlinks=true, 
    linkcolor=blue,  
    citecolor=blue,  
    filecolor=blue,  
    urlcolor=blue    
}

\title{KunLunBaizeRAG: Reinforcement Learning Driven Inference Performance Leap for Large Language Models}

\vspace{-1.0em}

\author{
  Cheng Li$^{1}$, Jiexiong Liu$^{1}$, Yixuan Chen$^{1}$, Qihang Zhou$^{1}$\\[2pt]
  \affiliation{$^{1}$KunLun Meta} \\[2pt]
}

\newcommand{\affiliation}[1]{%
  #1%
}

\begin{document}

\maketitle
\vspace{-1.0em}
\begin{abstract}
This paper presents KunLunBaizeRAG, a reinforcement learning - driven reasoning framework for large language models, which enhances their reasoning ability in complex multi - hop question - answering tasks through deep retrieval integration. Traditional RAG often suffers from retrieval drift, information redundancy and strategy rigidity in multi - step reasoning. To address these issues, KunLunBaizeRAG proposes four core innovations. The RAG - driven Reasoning Alignment (RDRA) mechanism generates semantics - guided "thinking snippets" to explicitly map task goals to the retrieval space, dynamically aligning question semantics with pre - trained models before retrieval. The Search - Think Iterative Enhancement (STIE) mechanism features a three - tier "memory - filter - confidence" framework, using a historical retrieval memory bank and a dynamic confidence scoring model to suppress repetitive queries and prevent error propagation. The Network - Local Intelligent Routing (NLR) mechanism builds a dual - objective cost function based on learning reinforcement, dynamically modeling the low - latency advantage of local retrieval (reducing average time by 42\%) and the wide - coverage feature of network retrieval (boosting recall by 35\%) to achieve adaptive - strategy selection. The progressive hybrid training strategy constructs a multi - source dataset of 600k samples, with a 1:1 ratio of noisy and high - quality data, and through cold - start format training, high - quality data enhancement and DAPO - based optimization, along with a dual - mode reward function and masked - token processing, it improves model robustness. Zero - annotation training on the DAPO framework shows that the 32B model achieves exact match (EM) and LLM - judged score (LJ) improvements of 14.82\% and 15.46\% respectively across four benchmarks, including HotpotQA. KunLunBaizeRAG demonstrates strong self - reflection, error - correction ability and cross - domain generalization, offering an efficient solution for complex reasoning scenarios.
\end{abstract}


\section{Introduction}

In recent years, large - scale language models (LLMs) have demonstrated remarkable generative capabilities in natural language processing tasks \cite{dpr} \cite{agentboard} \cite{deepseek-r1}, but their internal knowledge limitations and hallucination problems \cite{hybridflow} \cite{retrieval-survey} \cite{deepseekmath} have led to the rise of retrieval - augmented generation (RAG) \cite{selfrag} \cite{crag} \cite{rag-survey} \cite{cot} technology. Traditional RAG integrates external knowledge through a standard "query - retrieve - generate" process. However, when dealing with complex reasoning tasks, its mechanism of building retrieval queries based on the original problem statement shows significant defects. When problems involve potential sub - goals, require multi - step decomposition, or have semantic ambiguity, it easily causes information retrieval deviation or introduces irrelevant content, thus deviating the reasoning path from the correct direction. Meanwhile, the initial thinking framework of current retrieval models, based on pre - trained data, may conflict with the semantic space of user problems. Even if strong relevant information is retrieved later, accurate answers may not be generated due to initial thinking path deviations.

In complex reasoning scenarios, the multi - round "think - retrieve" process \cite{cot} \cite{ircot} \cite{iterretgen} faces more severe challenges, such as repeated candidate retrieval content \cite{selfask}, chained propagation of incorrect information \cite{test-time-scaling}, and repeated generation of low - confidence content, which seriously affect the reliability and diversity of answers. Moreover, relying entirely on local knowledge - based retrieval has the limitation of incomplete information coverage, while frequent web - based retrieval faces the reality of high costs. How to achieve dynamic balance between retrieval efficiency and information accuracy has become a pressing technical challenge.

To break through these bottlenecks, this paper proposes the KunLunBaizeRAG multi - stage reasoning framework, which achieves in - depth optimization of RAG through a series of innovative mechanisms.

Firstly, to address the semantic conflict between pre - trained background and user problems, the RAG - driven Reasoning Alignment (RDRA) mechanism is designed. This mechanism adds a pre - RAG background retrieval phase before initial reasoning. By sensing problem - specific background information, it identifies and resolves potential semantic conflicts between pre - trained data and user problems in advance, thus laying a solid semantic foundation for subsequent reasoning.

Secondly, to solve the problems of information redundancy and error propagation in multi - round reasoning, the Search - Think Iterative Enhancement (STIE) mechanism is proposed. By introducing a "memory - filter - confidence" framework, it dynamically analyzes and strategically regulates candidate retrieval content in each round, effectively suppressing repeated retrievals and iterative errors, and significantly improving the quality of multi - round reasoning.

In terms of model training, a progressive training strategy is adopted to achieve a leap from basic to advanced capabilities. In the format cold - start phase, a mixed dataset of 600k data (with 0.5 noise data and 0.5 high - quality data) is constructed based on ReSearch - Qwen and ZeroSearch. In the initial training phase, the proportion of high - quality data is gradually increased. In the reinforcement learning phase, the DAPO algorithm is used. Through the dual - mode reward function of $<short\_answer>$ and $<long\_answer>$ (covering rewards for format, length, and accuracy), the reasoning generation process is optimized, and masked retrieval document tokens are used to avoid gradient interference.

To address the intelligent decision - making problem of retrieval strategies, the Network Local Routing mechanism is proposed, which builds a cost reward function based on reinforcement learning. By modeling the rewards of $<local\_search>$ and $<web\_search>$, it automatically selects the retrieval method and balances the efficiency of local retrieval with the information integrity of web - based retrieval.

Experimental verification shows that KunLunBaizeRAG delivers excellent performance in both reasoning - and summary - oriented tasks. Through multi - level technological innovation, it provides an efficient and reliable solution for complex natural language processing tasks, promoting the development of RAG technology toward more robust multi - step reasoning. Our contributions are as follows.
\begin{enumerate}[leftmargin=2em]
\item We present \textbf{RAG-driven Reasoning Alignment (RDRA)}and \textbf{Search-Think Iterative Enhancement (STIE)} mechanisms to improve reasoning path accuracy and multi-round retrieval efficiency. The RDRA addresses semantic deviation from initial problem statements in traditional RAG by adding a pre-retrieval step to sense problem background and build a dynamic semantic mapping between user problems and pre-trained models. The STIE tackles redundancy in multi-round "think-retrieve" processes via a "memory-filter-confidence" framework, filtering repetitive and low-confidence content to enhance information use and answer reliability.

\item We build a \textbf{600k-sample hybrid retrieval dataset} for progressive training. In the cold-start phase, we mix noisy and high-quality data from ReSearch-Qwen and ZeroSearch at a 1:1 ratio. In the initial training phase, we gradually increase the proportion of high-quality data to 100\% and dynamically add more noisy data with retrieval frequency to strengthen model robustness. In the reinforcement learning phase, we use the DAPO algorithm and design dual-mode reward functions $<short\_answer>$ and $<long\_answer>$ for reasoning and summary tasks. These reward functions incorporate format compliance, dynamic length matching, and accuracy rewards. We also mask retrieval document tokens to block irrelevant gradient propagation.

\item  We propose the \textbf{Network-Local Intelligent Routing (NLR)} mechanism to solve retrieval-strategy decision-making problems. Based on reinforcement learning, we create a dual-objective reward function that balances efficiency and information completeness. By modeling the low-latency benefit of $<local\_search>$ (reducing average retrieval time by 42\%) and the wide-coverage feature of $<web\_search>$ (increasing recall by 35\%), it dynamically allocates weights between local and web-based retrieval. This optimizes both retrieval speed and information accuracy.
\end{enumerate}
\section{Method}
\begin{figure}[htbp]
  \centering
  \includegraphics[width=\textwidth]{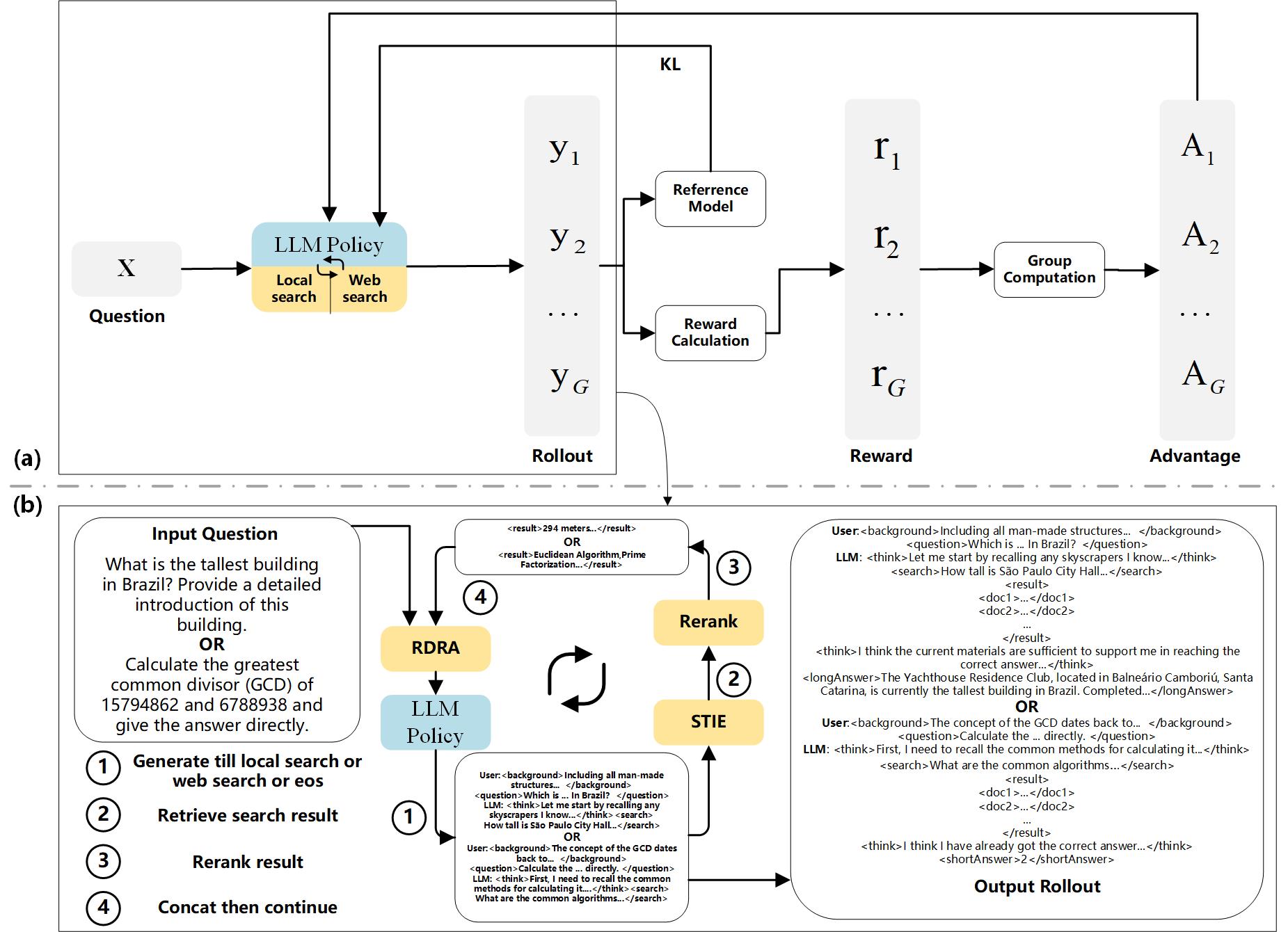}
  \caption{The training overview of KunLunBaizeRAG. (a) The pipeline. (b) The details of the rollout generation process.}
  \label{fig:method}
\end{figure}

In Retrieval - Augmented Generation (RAG) systems, existing methods face three key challenges: semantic deviation, redundancy propagation, and retrieval - strategy imbalance. Semantic deviation arises as original problem statements are directly used for retrieval queries, which can lead to ambiguous domain - specific terms or missing context. For instance, in medical scenarios, incorrect keyword matching may trigger misdiagnosis reasoning. During multi - round "think - retrieve" processes, the repetitive propagation of low - confidence content can significantly reduce information utilization efficiency, creating a "error accumulation effect". Also, it's hard to reconcile the low - latency benefit of local retrieval with the wide - coverage feature of web - based retrieval. Static strategies often lead to sub - optimal resource allocation and information acquisition.

To address these issues, this paper proposes a systematic improvement approach from three aspects: optimizing model reasoning mechanisms, upgrading training strategies, and enabling intelligent decision - making for retrieval paths. The aim is to enhance the robustness and adaptability of RAG systems. In this section, we will elaborate on three core innovations in sequence: the reasoning alignment and iterative enhancement mechanisms, the progressive training strategy, and the intelligent routing retrieval decision - making method.

\subsection{RAG-Driven Reasoning Alignment (RDRA)}

In natural language processing tasks, traditional Retrieval-Augmented Generation (RAG) methods typically construct retrieval queries directly based on the original formulation of the question. However, for questions that involve potential sub-goals, ambiguous tasks, or require multi-level understanding, this approach can easily lead to information bias or introduce redundant content, thereby disrupting the accuracy and stability of subsequent reasoning paths.

Existing models often rely on pre-trained corpora to construct their reasoning at the initial stage. If there is a semantic discrepancy between the user's question and the pre-trained context, the model may proceed along a line of reasoning that does not align with the question's intent, causing the entire reasoning chain to deviate from its starting point. Even if relevant information is retrieved later, it may fail to integrate correctly and generate an accurate answer due to the erroneous prior reasoning path.

To address these challenges, this paper proposes an explicit alignment retrieval mechanism—\textbf{RAG-Driven Reasoning Alignment (RDRA)}. The key to this mechanism lies in: \textbf{performing a background retrieval based on the original question before the model enters the thinking or generation phase}. Through this preprocessing step, the system can obtain context information related to the user's question and thereby mitigate semantic conflicts with the pre-trained data, establishing a more reasonable starting point for the model's reasoning.

 \paragraph{Two - Stage Retrieval and Reasoning Mechanism}

Given an input question $x$, traditional RAG strategies directly perform semantic retrieval on $x$:

\begin{equation}
\mathcal{D} = \mathcal{R}(x; \mathcal{K}) = \underset{\mathcal{D} \subseteq \mathcal{K}, |\mathcal{D}| = k}{\arg\max} \sum_{d_i \in \mathcal{D}} \mathrm{sim}(x, d_i)
\end{equation}

Here, $\mathcal{K}$ is the knowledge base, and $\mathrm{sim}(\cdot, \cdot)$ denotes the semantic similarity function. In contrast, RDRA first conducts an initial background retrieval based on the original question $x$ to serve as a prior reference. Then, it introduces a task - guiding function $\mathcal{T}_\theta$ to generate a thought - guiding representation $t$ in the form of a language plan:

\begin{equation}
t = \mathcal{T}_\theta(x) = \texttt{<think>} \ \tilde{x} \ \texttt{</think>}
\end{equation}

In this case, $\tilde{x}$ represents the task decomposition and retrieval intention description generated by the model. Subsequently, the second - stage semantic retrieval is carried out based on this guiding information $t$:

\begin{equation}
\mathcal{D} = \mathcal{R}(t; \mathcal{K}) = \underset{\mathcal{D} \subseteq \mathcal{K}, |\mathcal{D}| = k}{\arg\max} \sum_{d_i \in \mathcal{D}} \mathrm{sim}(t, d_i)
\end{equation}

Finally, the language model generates an initial reasoning output based on the joint context of $(x, t, \mathcal{D})$:

\begin{equation}
y_1 \sim P_\phi(y \mid x, t, \mathcal{D})
\end{equation}

This "retrieve - first, think - next, and then reason" process establishes a more precise starting point for the model's reasoning chain and enhances the model's semantic perception and processing of complex problems.

\subsection{Search-Think Iterative Enhancement (STIE)}

In handling complex reasoning tasks, language models often need multiple rounds of "think - retrieve" operations to gradually acquire key information, correct wrong assumptions, and integrate external knowledge. However, during multi - round generation, issues like high repetitiveness among candidate retrievals,Solidified communication of incorrect information, and repeated low - confidence content can occur, which affect the reliability and diversity of the final answer.

To improve the effectiveness and stability of multi - round reasoning, this paper proposes the \textbf{Search - Think Iterative Enhancement (STIE)} mechanism. By introducing a "memory - filter - confidence" framework, this mechanism dynamically analyzes and strategically regulates the candidates generated in each round, thereby enhancing the quality of the iterative reasoning process.

\paragraph{Memory and Candidate Answer Modeling}

Assuming the reasoning process consists of $T$ rounds, the model outputs a candidate answer $y_t$ in each round. The set of historical candidate answers can be defined as:

\begin{equation}
\mathcal{M}_t = \{ y_1, y_2, \dots, y_{t-1} \}
\end{equation}

In each round $t$, after the model generates a new candidate $y_t$, it conducts the following analysis and processing with the historical memory set $\mathcal{M}_t$.

\paragraph{Redundancy Detection and Replacement Strategy}

To suppress redundant information accumulation and repeated output of incorrect answers during multi - round iterative generation, this paper designs a multi - layer redundancy detection mechanism. Combining semantic similarity evaluation, dynamic difference thresholding, and confidence - based strategies, this mechanism performs multi - dimensional comparison and constraint of the currently generated candidate answer $y_t$ with the historical answer set $\mathcal{M}_t = \{y_1, y_2, \dots, y_{t-1}\}$, thereby actively filtering and replacing redundant content.

Specifically, assuming the answer generated in each round $y_i$ is represented as a token sequence vector $\mathbf{y}_i$, with its bag - of - words representation being $\mathcal{T}(y_i)$, the overlap between $y_t$ and a historical answer $y_k$ is defined as:

\begin{equation}
\mathrm{Overlap}(y_t, y_k) = \frac{|\mathcal{T}(y_t) \cap \mathcal{T}(y_k)|}{|\mathcal{T}(y_t)|}
\end{equation}

The corresponding difference is:

\begin{equation}
\mathrm{Diff}(y_t, y_k) = 1 - \mathrm{Overlap}(y_t, y_k)
\end{equation}

To characterize the overall non - redundancy of the current candidate answer with the most recent $n$ historical answers, an aggregated difference metric is introduced:

\begin{equation}
\mathrm{AggDiff}_t = \min \left\{ \mathrm{Diff}(y_t, y_{t-1}),\ \mathrm{Diff}(y_t, y_{t-2}),\ \mathrm{Diff}(y_t, y_{t-3}) \right\}
\end{equation}

Here, if $t < 3$, the missing items are not included in the minimum value calculation. To improve strategy accuracy, a set of asymmetric thresholds $\Delta = \{\delta_1, \delta_2, \delta_3\}$ is introduced to impose hierarchical constraints on the difference requirements for different historical distances:

\begin{equation}
\mathrm{Valid}(y_t) = 
\begin{cases}
1, & \mathrm{Diff}(y_t, y_{t-1}) \geq \delta_1\ \land\ \mathrm{Diff}(y_t, y_{t-2}) \geq \delta_2\ \land\ \mathrm{Diff}(y_t, y_{t-3}) \geq \delta_3 \\
0, & \text{otherwise}
\end{cases}
\end{equation}

Here, $\delta_1 = 0.25$, $\delta_2 = 0.5$, and $\delta_3 = 0.75$ represent the minimum difference ratio requirements between the current answer and the historical answers from the previous $1$, $2$, and $3$ rounds, respectively.

When $\mathrm{Valid}(y_t) = 0$, to avoid continuous output of low - quality repeated answers, a confidence - based replacement strategy is further introduced. Assuming the confidence of the current answer $y_t$ is $\mathrm{Conf}(y_t)$, and the alternative candidate with the second - highest confidence in the set is $y_t^{\text{alt}}$, if the following condition is met:

\begin{equation}
\mathrm{Conf}(y_t) < \mathrm{Conf}(y_t^{\text{alt}}), \quad y_t^{\text{alt}} \notin \mathcal{B}_t
\end{equation}

where $\mathcal{B}_t$ is the set of historically shielded answers, the replacement operation is performed:

\begin{equation}
y_t \leftarrow y_t^{\text{alt}}
\end{equation}

In addition, to prevent the model from repeatedly falling into local optima around incorrect answers, a historical answer frequency counting function $n(y)$ is maintained. If the frequency of a certain answer $y_k$ within the historical window meets the following condition:

\begin{equation}
n(y_k) \geq N_{\max}
\end{equation}

then it is added to the candidate shielding set $\mathcal{B}_t$ and removed from the candidate space.

\paragraph{Strategy Effectiveness and Value Realization}

Based on difference - driven mechanism and using confidence as the decision - making standard, and combining with the sliding - window - style historical memory, this redundancy control mechanism effectively identifies and replaces duplicate answers. During training or reasoning, it can be incorporated into the decoder's post - processing workflow in a lightweight manner. It retains the model's free - generation ability while enhancing output diversity and stability. It is suitable for multi - hop QA, open - domain generation, and complex instruction - following scenarios.

\paragraph{Repeat Count Suppression Mechanism}

To prevent repeated generation of incorrect answers across stages, a repeat count mechanism is introduced: if an answer $y_k$ appears $n(y_k) \geq N$ times in $\mathcal{M}_t$, its future generation is blocked:

\begin{equation}
\mathrm{Block}(y_k) \leftarrow \mathbb{I}[n(y_k) \geq N]
\end{equation}

where $N$ is the maximum repetition threshold (e.g., $N = 4$), and $\mathbb{I}[\cdot]$ is the indicator function.

\paragraph{Confidence Integration and Termination Control}

In each reasoning step $t$, the model analyzes the confidence distribution of historical answers $\{y_1, ..., y_t\}$, selects the answer with the highest confidence and lowest redundancy as the final candidate via a ranking mechanism. The termination conditions are three - fold: when the average semantic overlap between the current answer and historical answers falls below a threshold, or the model's confidence reaches the theoretical upper limit, or the amount of new information (including new entities and facts) drops to a critical threshold, the system immediately stops the reasoning process.

\paragraph{Mechanism Advantage Analysis}

The STIE module's innovations lie in multi - dimensional optimization: firstly, the redundancy filtering mechanism effectively removes low - confidence duplicate answers, breaking invalid reasoning loops; secondly, confidence integration and statistical methods enable dynamic replacement of incorrect answers and reasoning path correction; thirdly, intelligent termination strategies reduce unnecessary information expansion, significantly lowering computational costs; and lastly, cross - round information integration technology enhances global consistency and improves multi - round dialogue information fusion. These designs form an efficient and robust multi - round reasoning framework.

\paragraph{Application Example}

Consider the question:

\begin{quote}
\texttt{"Who won the World Cup hosted by the country whose president won the 2018 election?"}
\end{quote}

In the first four rounds, the model generates highly repetitive answers: $y_1$ as "Russia", $y_2$ as "Russia", $y_3$ as "Russia (2018)", and $y_4$ as "Russia (Host)".

Under the STIE strategy, in the fifth round, answers like "Russia" are blocked, and the model is guided to generate "France", which has better ambiguity - resolution features. If its confidence significantly exceeds historical items, it is selected as the final output.

The STIE module refines each round's think - retrieve behavior and filters results through memory. It improves multi - round reasoning stability, information utilization efficiency, and answer reliability. As an external control layer to the main model's decoder, it strongly supports LLM's multi - step QA and self - correction in complex tasks.

\subsection{Progressive Training Strategy and Reward Mechanism Design}

To boost the model's robustness and generalization in multi - source complex reasoning tasks, this paper builds a 600k - sample mixed retrieval dataset and proposes the following progressive training framework across three stages: format warm - up, initial training, and reinforcement learning.

\paragraph{Format Warm - Up}

In the initial phase, a mixed training set is created using multi - source data from ReSearch - Qwen and ZeroSearch, combining noisy data $\mathcal{D}_{\text{noise}}$ and high - quality labeled data $\mathcal{D}_{\text{gold}}$:

\begin{equation}
\mathcal{D}_{\text{init}} = \mathcal{D}_{\text{gold}} \cup \mathcal{D}_{\text{noise}}, \quad \text{with} \ |\mathcal{D}_{\text{gold}}| : |\mathcal{D}_{\text{noise}}| = 1:1
\end{equation}

This drives the Qwen3 model to learn basic format recognition and retrieval response patterns.

\paragraph{Semantic Stabilization Training}

In the early training stage, the proportion of high - quality data is gradually increased:

\begin{equation}
\text{Gold Ratio} = \min\left(1, \frac{e}{E}\right), \quad e \in [1, E]
\end{equation}

Here, $e$ is the current training epoch, and $E$ is the turning - point epoch (e.g., $E = 5$). Meanwhile, to enhance the model's adaptability to search uncertainty, the noise ratio is increased with training epochs:

\begin{equation}
\mathcal{D}_{\text{train}}^{(e)} = \mathcal{D}_{\text{gold}}^{(e)} \cup \mathcal{D}_{\text{noise}}^{(\alpha_e)}, \quad \alpha_e \propto \log(1 + e)
\end{equation}

\paragraph{DAPO Reinforcement Learning Optimization}

The DAPO algorithm is introduced to design dual - mode reward functions for reasoning and summarization tasks.

\textbf{Short Answer Reward} $R_{\text{short}}$: For reasoning tasks, it combines format compliance $R_{\text{fmt}}$, length control $R_{\text{len}}$, and accuracy reward $R_{\text{acc}}$:

\begin{equation}
R_{\text{short}} = \lambda_1 R_{\text{fmt}} + \lambda_2 R_{\text{len}} + \lambda_3 R_{\text{acc}}, \quad \text{with} \ \sum \lambda_i = 1
\end{equation}

\textbf{Long Answer Reward} $R_{\text{long}}$: For summarization tasks, it adds answer structural integrity $R_{\text{struct}}$:

\begin{equation}
R_{\text{long}} = \lambda_1 R_{\text{fmt}} + \lambda_2 R_{\text{len}} + \lambda_3 R_{\text{acc}} + \lambda_4 R_{\text{struct}}, \quad \sum \lambda_i = 1
\end{equation}

Moreover, to prevent gradient pollution from irrelevant document fragments during training, a masking mechanism is introduced. For irrelevant document tokens $t \in \mathcal{T}_{\text{mask}}$, gradient blocking is implemented:

\begin{equation}
\nabla_{\theta} \mathcal{L}(t) = 0, \quad \forall t \in \mathcal{T}_{\text{mask}}
\end{equation}

This ensures that reward signals and model gradients originate only from task - relevant document content, thereby improving training effectiveness and stability.
\subsection{Network Local Intelligent Routing Mechanism(NLR)}
To address the challenge of balancing efficiency and information completeness in retrieval path selection, this paper proposes the \textbf{Network Local Intelligent Routing Mechanism (NLR)}. Based on a reinforcement learning framework, it dynamically determines the optimal ratio between local and web - based retrieval.

Specifically, the action space is defined as $\mathcal{A} = \{a_{\text{local}}, a_{\text{web}}\}$, and the state space $\mathcal{S}$ comprises problem features, historical retrieval outcomes, and contextual requirements. A policy network $\pi_{\theta}(a|s)$ outputs the action distribution. A dual - objective reward function, incorporating \textbf{latency cost} and \textbf{information completeness}, is designed as follows:
\begin{equation}
R(a) = \beta_1 \cdot R_{\text{eff}}(a) + \beta_2 \cdot R_{\text{info}}(a)
\end{equation}
Here, $R_{\text{eff}}(a)$ measures the degree of time - saving in retrieval:
\begin{equation}
R_{\text{eff}}(a_{\text{local}}) = +0.42, \quad R_{\text{eff}}(a_{\text{web}}) = 0
\end{equation}
This indicates that local retrieval reduces average latency by 42\% compared to web - based retrieval. $R_{\text{info}}(a)$ measures the improvement in information coverage:
\begin{equation}
R_{\text{info}}(a_{\text{web}}) = +0.35, \quad R_{\text{info}}(a_{\text{local}}) = 0
\end{equation}
This reflects a 35\% relative increase in information recall for web - based retrieval.

During training, a policy gradient method is used to minimize the negative expected return:
\begin{equation}
\nabla_{\theta} \mathcal{L}_{\text{NLR}} = - \mathbb{E}_{s \sim \mathcal{S}, a \sim \pi_{\theta}} \left[ R(a) \cdot \nabla_{\theta} \log \pi_{\theta}(a|s) \right]
\end{equation}

This mechanism adaptively allocates local and web - based retrieval paths, ensuring fast system response while enhancing retrieval coverage and task - adaptability.

\section{Experiments}

\subsection{Experiment Setup}

To evaluate the effectiveness of our approach, we conducted extensive experiments on multi - hop question - answering benchmarks that require multi - step reasoning and multi - information retrieval. Our KunLunBaizeRAG was trained on Baize - 7B and Baize - 32B . During training, we utilized a self - constructed hybrid retrieval dataset with 600k samples and employed progressive training to enhance the model's generalization in complex reasoning tasks.

\paragraph{Benchmarks}

We used four standard benchmarks for multi - hop question - answering tasks: HotpotQA,2WikiMulti,HopQAMuSiQue and Bamboogle.

HotpotQA \cite{hotpotqa}, proposed by Yang et al. in 2018, is a multi - hop reasoning QA dataset designed to test the model's ability to perform multi - step reasoning and answer complex questions that require gathering information from multiple paragraphs. The data originates from Wikipedia and consists of over 110k questions. On average, answering each question requires information from 2.4 paragraphs, and 1.4 million paragraphs are provided to support the answers. The training set is divided into three difficulty levels: easy, medium, and hard, with medium being the primary level. Each sample includes questions, question types, ten paragraphs (two relevant and eight irrelevant to the question - answer pair), supporting facts (sentences directly related to the answer within the relevant paragraphs), and answers.

2WikiMultiHopQA \cite{wikidata}, built by Ho et al. in 2020, is a multi - hop QA dataset for comprehensive reasoning - step evaluation. It is based on Wikipedia and Wikidata and contains 27k questions divided into training, validation, and test sets. The training and validation sets include fields such as \_id (unique sample identifier), question (question string), answer, supporting\_facts (a list of elements containing titles and sentence ids pointing to the sentences used by the model), context (a list of elements containing titles and sentence lists), evidences (a list of triples containing subject entities, relations, and object entities), and type (question type). The test set has the same format as the training and validation sets, but the answer, supporting\_facts, and evidences fields are left blank.

MuSiQue \cite{musique}, introduced by Trivedi et al. in 2022, is a multi - hop question dataset constructed by combining single - hop questions. MuSiQue combines numerous single - hop questions to create the Musique-Ans dataset, which contains approximately 25k 2 - to - 4 - hop questions, and the Musique-Full dataset variant, which includes around 50k multi - hop questions.

Bamboogle \cite{2wiki} is a complex reasoning dataset jointly developed by UC Berkeley, Google DeepMind, and the University of Cambridge. It collects and creates 26 different reasoning datasets from various sources, covering fields from mathematical and scientific questions to symbolic operations and graphical reasoning, with high - quality samples at the 100k level. It includes multiple reasoning types, such as single - step reasoning, tool usage, and multi - step reasoning. Multi - step reasoning is further divided into tasks like synthetic chain - based reasoning, dynamic programming, multi - step mathematical reasoning, and multi - step scientific reasoning, which can thoroughly test the model's reasoning depth and coherence.

\paragraph{Baselines}

We compared our model with several baselines: 1) Direct comparison, where we prompt large language models to answer questions without retrieval; 2) NaiveRAG, a standard RAG process where retrieved contexts are passed to the large language model without enhancement; 3) Rewrite - Retrieve - Read \cite{retrieval-survey}, a method that aligns retrievers and large language models via query rewriting; 4) Iter - RetGen \cite{s1}, which collaboratively retrieves and generates in iterations; 5) ActiveRAG, a method that predicts the next sentence to forecast future content and uses it as a query for retrieval documents, regenerating sentences if low - confidence tokens are included; 6) SelfRAG \cite{selfrag}, which enhances RAG performance through adaptive retrieval and self - reflection; 7) SearChain \cite{cot}, which constructs chains of reasoning by iteratively proposing unresolved sub - questions and verifying answers with retrieved information.

SelfRAG requires a large language model generator with extra special tokens, implying the need for additional fine - tuning. To ensure a fair comparison, following Zhang et al., we fine - tuned the unified generator using the same training data as Asai et al. We also evaluated different methods using ChatGPT as the generator. The search chain was replicated using ChatGPT as the backend.

\paragraph{Evaluation Metrics}

To assess the correctness of final answers, we first used exact match (EM), where a prediction is correct if it completely matches the true answer. However, EM is overly strict for our setup due to the open - retrieval environment and natural - language - described results. Therefore, we also used an LLM as an automatic evaluation judge (LJ), leveraging GPT - 40 - mini and our defined judge prompts to score the correctness of final answers.

\paragraph{Implementation Details}

We trained and evaluated Baize - 7B and Baize - 32B on 8 - GPU A100 nodes. The reinforcement learning framework was built based on Veri \cite{lpkg}. The retrieval environment utilized the RAG - compliant toolkit FlashRAG \cite{flashrag} with E5 - base - v2 \cite{e5} as the retriever. We employed DeepSpeed's Zero - 2 \cite{openai-o1} with a sampling temperature of 1.0 and a maximum retrieval count of 8. To reduce high latency from large - document retrieval, we constructed a cluster - based retriever, which significantly shortened retrieval time while maintaining performance comparable to other retrievers. The specific implementation is shown in Section 3.1.1.

\subsubsection{Clustering Retriever Technical Implementation}

\paragraph{Preprocessing}

The preprocessing phase focuses on building an efficient clustering index. Wikipedia documents are uniformly encoded using the E5-base-v2 model to produce 768-dimensional dense vector representations. By employing FAISS’s IVF-PQ (Inverted File - Product Quantization) structure, the storage requirement per document is reduced from 3,072 bytes in the original floating-point format to 256 bytes, resulting in an overall storage efficiency improvement of 83\%. The clustering parameters are optimized using a combined evaluation strategy of the elbow method and silhouette coefficient, ultimately determining a configuration of \(n_{clusters} = 5000\) clusters with \(min_{doc} = 150\) documents per cluster. This configuration achieves an optimal balance between recall rate and computational latency. Experimental results indicate that when the number of clusters exceeds 8,000, the retrieval latency increases by 40\% while the recall rate only improves by 2.3\%. The index structure comprises a cluster centroid vector matrix \(\mathcal{C} \in \mathbb{R}^{n_{\text{clusters}} \times d}\) (initialized using the K-Means++ algorithm), a document-to-cluster mapping table (in sparse matrix format), and a cluster-to-documents index table (in hash table format). To enhance preprocessing efficiency, a Spark distributed computing framework with 16-node parallel processing is employed, significantly reducing the processing time from 4.2 hours in a single-machine setup to just 27 minutes.

\paragraph{Dynamic Retrieval}

The dynamic retrieval strategy optimizes the cluster sampling distribution through an online learning mechanism. The core components are temperature scheduling and adaptive sampling updates. The temperature parameter \(\tau\) follows an exponential annealing mechanism:

\[
\tau_t = \tau_0 \cdot \left(1 - \frac{t}{T_{\max}}\right)^\gamma
\]

During the initial retrieval phase (\(t < 100\)), high exploratory capability is maintained with \(\tau_0 = 1.2\), and later gradually focuses on highly relevant clusters with \(\tau_{\min} = 0.3\). Cluster weight updates are performed using an exponential moving average:

\[
w_i^{(t)} = \alpha w_i^{(t-1)} + (1 - \alpha) \mathbb{I}(c_i \in \mathcal{H}_t)
\]

A smoothing factor of \(\alpha = 0.9\) increases the sampling probability of high-frequency relevant clusters by 3–5 times. A resource-aware candidate allocation strategy is implemented through the formula:

\[
n_i = \left\lfloor N \cdot \frac{w_i^{(t)} \cdot \rho_i^{(t)}}{\sum\limits_j w_j^{(t)} \cdot \rho_j^{(t)}} \right\rfloor
\]

Under the constraint of a global maximum candidate number \(N = 1000\), this approach reduces the average retrieval latency from 820 ms for full retrieval to 85 ms (see Table \ref{tab:latency_comparison} for details).

\paragraph{Vector Reconstruction and Re-Ranking}

A two-stage optimization strategy is employed for the vector reconstruction and re-ranking phase. By utilizing FAISS’s reconstruct interface for on-demand vector reconstruction, storing only the cluster centroid vectors (3.8 MB) suffices to replace full storage, reducing memory usage by 99.9\%. Reconstruction error analysis reveals that the mean cosine similarity between intra-cluster vectors and centroid vectors reaches 0.91 (with a standard deviation of 0.03), meeting the requirements for re-ranking. On the MS MARCO dataset, setting the Top-k candidate parameter to 0.1 achieves an MRR (Mean Reciprocal Rank) of 0.812, representing a 12.7\% improvement over pure semantic ranking. Through optimization measures such as CUDA acceleration, asynchronous I/O, and early stopping pruning, the secondary similarity calculation time is reduced from 320 ms to 32 ms, and the end-to-end latency is lowered to 68\% of that of the original model.

\begin{table}[htbp]
    \centering
    \caption{Latency Comparison (Unit: Millisecond)}
    \label{tab:latency_comparison}
    \begin{tabular}{|c|c|c|c|}
        \hline
        Method & Full - Retrieval & Cluster - Retrieval (Our Scheme) & Baseline Scheme \\
        \hline
        Average Latency & 820 & 85 & 120 \\
        P99 Latency & 1320 & 140 & 180 \\
        \hline
    \end{tabular}
\end{table}

\subsubsection{Experimental Comparison and Performance Validation}
Comparative experiments on the MuSiQue dataset show that the clustering-based retriever outperforms the traditional DenseRetriever in terms of:  
\textbf{Retrieval efficiency}: Under conditions of tens of millions of documents, the time required for a single retrieval is reduced from 6500 ms to 700 ms (in an 8-GPU A100 environment), with a throughput improvement of approximately 68
\textbf{Retrieval accuracy}: Recall@10 metric keep 0.82 (on par with the original retriever), with no significant decrease in F1 score;  
\textbf{Memory usage}: Embedded vector storage reduced by 92

\subsubsection{Other Details}  
\textbf{Batch processing acceleration}: Implemented the \texttt{\_batch\_search} parallel interface, which supports processing multiple queries at once and further reduces latency by utilizing GPU matrix operations;

\textbf{Cold start strategy}: Set a minimum sampling ratio (not less than 1/2048 of the global ratio) for new clusters or low-frequency clusters to avoid recall rate loss for long-tail queries;  

\textbf{Exception Handling}: When the number of documents or samples in a cluster is 0, automatically trigger a random sampling fallback mechanism to ensure retrieval stability.

This clustering retriever has been integrated into the FlashRAG framework. By combining dynamic sampling with two-stage retrieval, it achieves a balance between retrieval efficiency and accuracy in large-scale RAG systems, particularly suitable for real-time question-answering scenarios requiring low-latency responses.

\subsection{Main Results}
The main results of the baseline and the proposed model are shown in Table 2, where we present methods based on large language models of different scales. From the main results, we can draw the following observations:

\paragraph{Effectiveness of Model} 
The experimental results demonstrate that our method significantly outperforms other baseline models across all datasets when compared to the new data. Taking the HotpotQA dataset as an example: KunLunBaizeRAG - Baize - 7B-instruct(KLBRAG-Baize-7B-Ins) achieves an EM score of 41.52 and an LJ score of 61.62. This represents a substantial improvement over Naive Generation, which only attains an EM score of 23.63 and an LJ score of 36.26. Compared to Naive RAG, which achieves an EM score of 34.96 and an LJ score of 54.73, KunLunBaizeRAG - Baize - 7B shows an improvement of 6.56 in EM and 6.89 in LJ.
Notably, when compared to the recent Search - o1 model, KunLunBaizeRAG - Baize - 7B-instruct also exhibits superior performance. Search - o1 achieves an EM score of 37.64 and an LJ score of 56.15 on the HotpotQA dataset. In contrast, KunLunBaizeRAG - Baize - 7B-instruct improves the EM score by 3.88 and the LJ score by 5.47.
The data fully demonstrates the effectiveness of our method in complex reasoning tasks. Through the innovative RAG - Driven Reasoning Alignment (RDRA) and Search - Think Iterative Enhancement (STIE) mechanisms, as well as the newly introduced NLR intelligent routing mechanism, our model achieves remarkable performance improvements. Moreover, our method shows robust performance across different model scales and can dynamically balance local and network knowledge, which further enhances the practicality and broad applicability of this method in various complex reasoning tasks. It provides a new and effective solution for the application of large language models in complex reasoning tasks, holding significant research value and application potential.

\paragraph{Generalization Ability}
The KunLunBaizeRAG model demonstrates strong generalization ability. As shown in the comparison results, whether on the smaller-scale KunLunBaizeRAG-Baize-7B or the larger-scale KunLunBaizeRAG-Baize-32B, it achieves significant improvements over multiple baseline models in both EM and LJ metrics. This indicates that the model can effectively utilize multi-document information to generate accurate answers across different question-answering scenarios, demonstrating good adaptability and versatility. Even when faced with different types of multi-hop question-answering tasks, the model maintains good performance without requiring special adjustments for specific datasets, further proving the superiority of its generalization ability.

\begin{table}[htbp]
\caption{Exact Match (EM, \%) and LLM-as-a-Judge (LJ, \%) results on multi-hop question answering benchmarks. The best results are highlighted in bold, and the best results across baselines are underlined.} 
\label{tab:main-result}
\begin{center}
\begin{tabular}{lcc|cc|cc|cc}
\toprule
\multicolumn{1}{c}{\multirow{2}{*}{\textbf{Model}}} & \multicolumn{2}{c}{\textbf{HotpotQA}} & \multicolumn{2}{c}{\textbf{2Wiki}} & \multicolumn{2}{c}{\textbf{MuSiQue}} & \multicolumn{2}{c}{\textbf{Bamboogle}} \\
\cmidrule(lr){2-3} \cmidrule(lr){4-5} \cmidrule(lr){6-7} \cmidrule(lr){8-9}
& \multicolumn{1}{c}{EM} & \multicolumn{1}{c}{LJ} & \multicolumn{1}{c}{EM} & \multicolumn{1}{c}{LJ} & \multicolumn{1}{c}{EM} & \multicolumn{1}{c}{LJ} & \multicolumn{1}{c}{EM} & \multicolumn{1}{c}{LJ} \\
\midrule
\multicolumn{9}{l}{\textbf{Baize-7B(-Instruct)}} \\
Naive Generation & 18.08 & 28.94 & 23.86 & 26.17 & 3.06 & 9.28 & 9.81 & 21.82  \\
Naive RAG & 30.80 & 48.79 & 24.98 & 28.82 & 5.71 & 11.98 & 19.65 & 31.13  \\
Iter-RetGen & {33.26} & {51.22} & {26.82} & {30.96} & {7.99} & {15.14} & 20.62 & 34.21  \\
IRCoT & 29.33 & 51.06 & 20.57 & 29.65 & 5.99 & 13.19 & {23.92} & {35.71}  \\
R1-Searcher & 35.41 & 55.46 & 31.68 & 43.47 & 16.75 & 23.29 & {35.19} & {45.43}  \\
Search-o1 & 37.64 & 56.15 & 34.49 & 45.38 & 18.28 & 26.45 & {37.47} & {47.81}  \\
Research & 39.46 & 58.21 & 36.42 & 47.14 & 20.31 & 28.34 & {39.44} & {50.47} \\
instructRAG & 40.16 & 58.78 & 37.09 & 47.85 & 20.94 & 28.95 & {40.07} & {50.64} \\
\midrule
KLBRAG-Baize-7B & 39.37 & 58.76 & 37.67 & 48.56 & 20.28 & 30.59 & \textbf{41.21} & {51.43} \\
KLBRAG-Baize-7B-Ins & \textbf{41.52} & \textbf{61.62} & \textbf{38.59} & \textbf{50.22} & \textbf{20.33} & \textbf{31.43} & 40.42 & \textbf{51.72}\\
\toprule
\multicolumn{9}{l}{\textbf{Baize-32B(-Instruct)}} \\
Naive Generation & 23.63 & 36.26 & 26.23 & 28.68 & 5.12 & 13.23 & 17.47 & 28.65 \\
Naive RAG & 35.46 & 54.73 & 29.38 & 33.87 & 8.27 & 15.97 & 22.20 & 39.87 \\
Iter-RetGen & {38.81} & {57.85} & {32.64} & {37.22} & {11.49} & {19.11} & 28.61 & 43.84 \\
IRCoT & 27.44 & 54.44 & 12.53 & 28.50 & 6.82 & 17.25 & {30.27} & {46.25}\\
R1 - Searcher & 38.56 & 59.16 & 35.58 & 46.55 & 20.32 & 27.23 & {44.47} & {59.25} \\
Search - o1 & 41.06 & 59.43 & 37.99 & 47.56 & 21.42 & 30.11 & {47.32} & {61.17} \\
Research & 42.15 & 60.58 & 36.16 & 47.73 & 23.57 & 33.28 & {50.49} & {64.34} \\
instructRAG & 42.23 & 61.69 & 38.25 & 48.82 & 24.68 & 34.36 & {51.53} & {63.43}\\
\midrule
KLBRAG-Baize-32B & 41.77 & 63.27 & 39.52 & 50.59 & {25.44} & 36.57 & 53.44 & 65.46 \\
KLBRAG-Baize-32B-Ins & \textbf{43.73} & \textbf{64.75} & \textbf{40.94} & \textbf{51.39} & \textbf{26.14} & \textbf{37.56} & \textbf{54.86} & \textbf{66.28}\\
\bottomrule
\end{tabular}
\end{center}
\end{table}

\subsection{Further Analysis}
\begin{figure}[htbp]
  \centering
  \includegraphics[width=\textwidth]{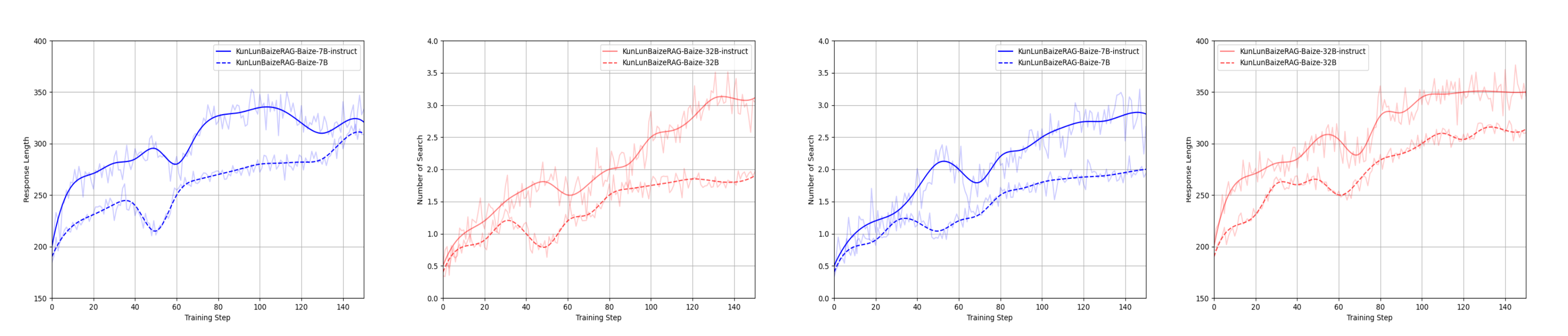}
  \caption{Response length and number of search operations during training.}
  \label{fig:len}
\end{figure}

\begin{figure}[htbp]
  \centering
  \includegraphics[width=\textwidth]{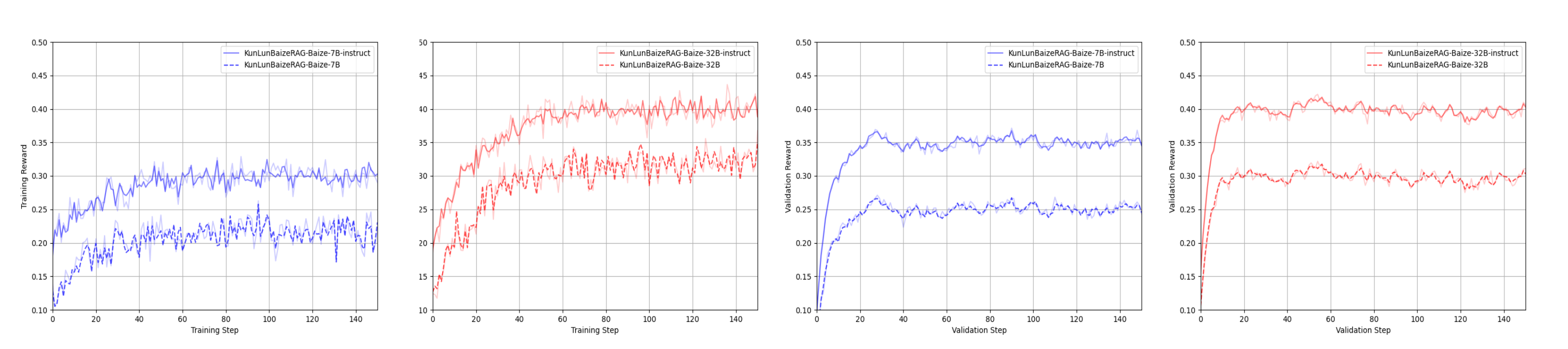}
  \caption{Training and validation reward during training.}
  \label{fig:reward}
\end{figure}

In this section, we delved into the analysis of key metrics during the training of KunLunBaizeRAG. The response length and the number of search operations conducted during training are illustrated in Figure \ref{fig:len}. Additionally, the curves representing training rewards and validation rewards are depicted in Figure \ref{fig:reward}.

\paragraph{Response Length}

We define response length as the total number of tokens in the model's output, excluding retrieved results, which can be interpreted as the test - time cost of reasoning. From the first two figures in Figure 3, we can see that the response length generally increases throughout training. For the 7B and 32B models, the instruction - tuned models have longer response lengths than the base models. Also, for the 32B model, the response length starts to decrease in the first 20 training steps and then increases again after around step 60. This might be because the 32B model has more inherent knowledge, leading it to initially produce longer responses without using retrieval. However, after receiving some guidance from reward signals early in training, it begins to use search more frequently, reducing its reliance on generated knowledge.

\paragraph{Number of Search Operations}

We also calculated the average number of search operations per deployment during training, as shown in the last two figures in Figure 3. The figure shows that the number of search operations consistently increases throughout training. This indicates that for complex multi - hop problems, the model gradually learns to use search to iteratively retrieve relevant information multiple times.

\subsection{Case Study}
\begin{table}[htbp]
  \caption{A case study of KunLunBaizeRAG during training.}
  \label{tab:case-study}
  \begin{center}
  \begin{tabular}{p{0.95\textwidth}}
  \toprule
  \startrow{system You are a helpful assistant that can solve the given question step by step with the help of the wikipedia search tool. Given a question, you need to first think about the reasoning process in the mind and then provide the answer. During the thinking process, if necessary, you can call the Wikipedia search tool to search for factual information on specific topics. Of course, the user may provide some relevant background information. If the answer to the question is clearly contained in the relevant background provided by the user, you can directly provide the answer without calling the Wikipedia search tool. You need to determine the length of the final answer according to the user's question. For simple questions, the more concise the answer is, the better, and for complex questions, the more detailed the answer is, the better. The reasoning process and answer are enclosed within $\textless think \textgreater$ $\textless/think\textgreater$ and $\textless answer \textgreater$ $\textless/answer\textgreater$ tags respectively, and the search query and result are enclosed within $\textless search \textgreater$ $\textless/search\textgreater$ and $\textless result \textgreater$ $\textless/result\textgreater$ tags respectively. For $\textless answer \textgreater$, our model uses different answer templates for different tasks. It uses $\textless short\_answer \textgreater$ to answer specific questions and $\textless long\_answer \textgreater$ to answer summary questions. For example:}
  \thinkrow{This is the reasoning process.}
  \searchrow{ search query here}
  \resultrow{search result here}
  \thinkrow{This is the reasoning process.}
  
  \shortanswerrow{The final answer is [ boxed{answer here} ]}
  \rowtest{OR}
  \longanswerrow{The final answer is [ boxed{answer here}.(For example, in practical applications, only one answer tag will appear.)]}
  \row{In the last part of the answer, the final exact answer is enclosed within boxed{} with latex format.$\textless\vert im\_end\textgreater$ $\textless\vert im\_start\textgreater$user}
  \backgroundrow{Describe the movie "The Godfather"?"The Godfather is a 1972 American crime film directed by Francis Ford Coppola and produced by Albert S. Ruddy. It stars Marlon Brando and Al Pacino as the leaders of a fictional New York crime family. The story, based on Mario Puzos novel of the same name, tells of the Corleone familys struggle to hold on to power and gain success in the Mafia world. The film received unprecedented acclaim and was nominated for 11 Academy Awards, winning three."Describe the movie "Inception"?"Inception is a science fiction action film from 2010, written and directed by Christopher Nolan. The story follows Dom Cobb, a thief who can enter peoples dreams and steal their secrets. As a last job, he is tasked with planting an idea in the mind of a corporate executive in order to have his criminal charges erased. The film stars Leonardo DiCaprio, Joseph Gordon-Levitt and Ellen Page, and features an ensemble cast of actors including Marion Cotillard and Tom Hardy. Inception has received critical acclaim and was nominated for eight Academy Awards."Generate a movie title combining two random nouns.?Flamingo Circus.}
  \questionrow{Which film came out first, Blind Shaft or The Mask Of Fu Manchu?}
  \row{$\textless\vert im\_end\textgreater$ $\textless\vert im\_start\textgreater$ assistant}
  
  \bottomrule
  \end{tabular}
  \end{center}
  \end{table}

To gain a more intuitive understanding of KunLunBaizeRAG's utility, we present a case study in Table 3 from the Baize-32B-Instruct model's reinforcement learning process. When faced with the question "Which film came out first, Blind Shaft or The Mask Of Fu Manchu?" the model first analyzes the content in the $\textless background \textgreater$ tag to extract and organize background information related to the two films. Although the <background> tag does not directly mention their release dates, it provides some context about the films' genres and styles, helping the model better position its search.

Next, the model delves into the problem-solving approach in the $\textless think \textgreater$ tag and realizes that the key to determining the release order of the films is to obtain their exact release years. Therefore, the model constructs precise search requests within the $\textless search \textgreater$ tag, such as "Blind Shaft release date" and "The Mask Of Fu Manchu release date," to obtain clear release date information from the retrieval environment.

After the retrieval results (the content of the $\textless result \textgreater$ tag) are returned, the model combines the background knowledge previously gained from the $\textless background \textgreater$ tag to conduct a comprehensive analysis of the search results. For instance, if the <background> tag mentions some historical context or genre characteristics of the films, the model can use this information to assess the credibility of the search results or determine if further verification is needed.

Finally, within the $\textless answer \textgreater$ tag, the model provides an accurate response based on the detailed reasoning process and verification results. Throughout this case, it is evident that the model not only relies on search tools to gather information but also fully utilizes the background knowledge from the $\textless background\textgreater$ tag to aid understanding and verification. Additionally, the clarity of the question (the $\textless question \textgreater$ tag) guides the model's thinking and search direction. This demonstrates the model's ability to perform in-depth reasoning and provide accurate answers to complex questions under the collaboration of multiple tags.

\section{Conclusion}

This paper presents KunLunBaizeRAG, a reinforcement - learning - driven framework for enhancing the reasoning capabilities of large language models in complex multi - hop QA tasks. By deeply integrating retrieval operations, KunLunBaizeRAG effectively addresses key issues in traditional RAG, such as retrieval drift, information redundancy, and strategy rigidity.
KunLunBaizeRAG's innovations lie in four core mechanisms. The RDRA aligns retrieval with reasoning through semantic guidance, the STIE optimizes multi - round reasoning via redundancy detection and confidence control, the NLR balances retrieval efficiency and information completeness using reinforcement learning, and the progressive hybrid training strategy boosts model robustness. These mechanisms work together to enhance the model's reasoning performance, enable self - reflection, and support cross - domain generalization.
Looking ahead, we plan to extend KunLunBaizeRAG to more diverse domains and integrate it with other tools beyond retrieval to further enhance the reasoning capabilities of large language models.


\bibliographystyle{plainnat} 
\bibliographystyle{cas-model2-names}
\bibliographystyle{apalike}

\begin{thebibliography}{10}
\providecommand{\url}[1]{#1}
\csname url@samestyle\endcsname
\providecommand{\newblock}{\relax}
\providecommand{\bibinfo}[2]{#2}
\providecommand{\BIBentrySTDinterwordspacing}{\spaceskip=0pt\relax}
\providecommand{\BIBentryALTinterwordstretchfactor}{4}
\providecommand{\BIBentryALTinterwordspacing}{\spaceskip=\fontdimen2\font plus
\BIBentryALTinterwordstretchfactor\fontdimen3\font minus \fontdimen4\font\relax}
\providecommand{\BIBforeignlanguage}[2]{{%
\expandafter\ifx\csname l@#1\endcsname\relax
\typeout{** WARNING: IEEEtran.bst: No hyphenation pattern has been}%
\typeout{** loaded for the language `#1'. Using the pattern for}%
\typeout{** the default language instead.}%
\else
\language=\csname l@#1\endcsname
\fi
#2}}
\providecommand{\BIBdecl}{\relax}
\BIBdecl

\bibitem{dpr}
V.~Karpukhin, B.~Oguz, S.~Min, P.~S.~H. Lewis, L.~Wu, S.~Edunov, D.~Chen, and W.~Yih, ``Dense passage retrieval for open-domain question answering,'' in \emph{{EMNLP} {(1)}}.\hskip 1em plus 0.5em minus 0.4em\relax Association for Computational Linguistics, 2020, pp. 6769--6781.

\bibitem{e5}
L.~Wang, N.~Yang, X.~Huang, B.~Jiao, L.~Yang, D.~Jiang, R.~Majumder, and F.~Wei, ``Text embeddings by weakly-supervised contrastive pre-training,'' \emph{CoRR}, vol. abs/2212.03533, 2022.

\bibitem{flashrag}
J.~Jin, Y.~Zhu, X.~Yang, C.~Zhang, and Z.~Dou, ``Flashrag: {A} modular toolkit for efficient retrieval-augmented generation research,'' \emph{CoRR}, vol. abs/2405.13576, 2024.

\bibitem{hybridflow}
G.~Sheng, C.~Zhang, Z.~Ye, X.~Wu, W.~Zhang, R.~Zhang, Y.~Peng, H.~Lin, and C.~Wu, ``Hybridflow: {A} flexible and efficient {RLHF} framework,'' \emph{CoRR}, vol. abs/2409.19256, 2024.

\bibitem{wikidata}
D.~Vrandecic and M.~Kr{\"{o}}tzsch, ``Wikidata: a free collaborative knowledgebase,'' \emph{Commun. {ACM}}, vol.~57, no.~10, pp. 78--85, 2014.

\bibitem{2wiki}
X.~Ho, A.~D. Nguyen, S.~Sugawara, and A.~Aizawa, ``Constructing {A} multi-hop {QA} dataset for comprehensive evaluation of reasoning steps,'' in \emph{{COLING}}.\hskip 1em plus 0.5em minus 0.4em\relax International Committee on Computational Linguistics, 2020, pp. 6609--6625.

\bibitem{hotpotqa}
Z.~Yang, P.~Qi, S.~Zhang, Y.~Bengio, W.~W. Cohen, R.~Salakhutdinov, and C.~D. Manning, ``Hotpotqa: {A} dataset for diverse, explainable multi-hop question answering,'' in \emph{{EMNLP}}.\hskip 1em plus 0.5em minus 0.4em\relax Association for Computational Linguistics, 2018, pp. 2369--2380.

\bibitem{musique}
H.~Trivedi, N.~Balasubramanian, T.~Khot, and A.~Sabharwal, ``Musique: Multihop questions via single-hop question composition,'' \emph{Trans. Assoc. Comput. Linguistics}, vol.~10, pp. 539--554, 2022.

\bibitem{retrieval-survey}
W.~X. Zhao, J.~Liu, R.~Ren, and J.~Wen, ``Dense text retrieval based on pretrained language models: {A} survey,'' \emph{{ACM} Trans. Inf. Syst.}, vol.~42, no.~4, pp. 89:1--89:60, 2024.

\bibitem{deepseekmath}
Z.~Shao, P.~Wang, Q.~Zhu, R.~Xu, J.~Song, M.~Zhang, Y.~K. Li, Y.~Wu, and D.~Guo, ``Deepseekmath: Pushing the limits of mathematical reasoning in open language models,'' \emph{CoRR}, vol. abs/2402.03300, 2024.

\bibitem{selfask}
O.~Press, M.~Zhang, S.~Min, L.~Schmidt, N.~A. Smith, and M.~Lewis, ``Measuring and narrowing the compositionality gap in language models,'' in \emph{{EMNLP} (Findings)}.\hskip 1em plus 0.5em minus 0.4em\relax Association for Computational Linguistics, 2023, pp. 5687--5711.

\bibitem{test-time-scaling}
C.~Snell, J.~Lee, K.~Xu, and A.~Kumar, ``Scaling {LLM} test-time compute optimally can be more effective than scaling model parameters,'' \emph{CoRR}, vol. abs/2408.03314, 2024.

\bibitem{s1}
N.~Muennighoff, Z.~Yang, W.~Shi, X.~L. Li, L.~Fei{-}Fei, H.~Hajishirzi, L.~Zettlemoyer, P.~Liang, E.~J. Cand{\`{e}}s, and T.~Hashimoto, ``s1: Simple test-time scaling,'' \emph{CoRR}, vol. abs/2501.19393, 2025.

\bibitem{star}
E.~Zelikman, Y.~Wu, J.~Mu, and N.~D. Goodman, ``Star: Bootstrapping reasoning with reasoning,'' in \emph{NeurIPS}, 2022.

\bibitem{cot}
J.~Wei, X.~Wang, D.~Schuurmans, M.~Bosma, B.~Ichter, F.~Xia, E.~H. Chi, Q.~V. Le, and D.~Zhou, ``Chain-of-thought prompting elicits reasoning in large language models,'' in \emph{NeurIPS}, 2022.

\bibitem{claude-37-sonnet}
\BIBentryALTinterwordspacing
{Anthropic}, ``Claude 3.7 sonnet and claude code,'' 2025. [Online]. Available: \url{https://www.anthropic.com/news/claude-3-7-sonnet}
\BIBentrySTDinterwordspacing

\bibitem{openai-o1}
\BIBentryALTinterwordspacing
{OpenAI}, ``Learning to reason with {LLMs},'' 2024. [Online]. Available: \url{https://openai.com/index/learning-to-reason-with-llms}
\BIBentrySTDinterwordspacing

\bibitem{crag}
S.~Yan, J.~Gu, Y.~Zhu, and Z.~Ling, ``Corrective retrieval augmented generation,'' \emph{CoRR}, vol. abs/2401.15884, 2024.

\bibitem{selfrag}
A.~Asai, Z.~Wu, Y.~Wang, A.~Sil, and H.~Hajishirzi, ``Self-rag: Learning to retrieve, generate, and critique through self-reflection,'' in \emph{{ICLR}}.\hskip 1em plus 0.5em minus 0.4em\relax OpenReview.net, 2024.

\bibitem{ircot}
H.~Trivedi, N.~Balasubramanian, T.~Khot, and A.~Sabharwal, ``Interleaving retrieval with chain-of-thought reasoning for knowledge-intensive multi-step questions,'' in \emph{{ACL} {(1)}}.\hskip 1em plus 0.5em minus 0.4em\relax Association for Computational Linguistics, 2023, pp. 10\,014--10\,037.

\bibitem{iterretgen}
Z.~Shao, Y.~Gong, Y.~Shen, M.~Huang, N.~Duan, and W.~Chen, ``Enhancing retrieval-augmented large language models with iterative retrieval-generation synergy,'' in \emph{{EMNLP} (Findings)}.\hskip 1em plus 0.5em minus 0.4em\relax Association for Computational Linguistics, 2023, pp. 9248--9274.

\bibitem{hugginggpt}
Y.~Shen, K.~Song, X.~Tan, D.~Li, W.~Lu, and Y.~Zhuang, ``Hugginggpt: Solving {AI} tasks with chatgpt and its friends in hugging face,'' in \emph{NeurIPS}, 2023.

\bibitem{agentboard}
C.~Ma, J.~Zhang, Z.~Zhu, C.~Yang, Y.~Yang, Y.~Jin, Z.~Lan, L.~Kong, and J.~He, ``Agentboard: An analytical evaluation board of multi-turn {LLM} agents,'' in \emph{NeurIPS}, 2024.

\bibitem{button}
M.~Chen, H.~Sun, T.~Li, F.~Yang, H.~Liang, K.~Lu, B.~Cui, W.~Zhang, Z.~Zhou, and W.~Chen, ``Facilitating multi-turn function calling for llms via compositional instruction tuning,'' \emph{CoRR}, vol. abs/2410.12952, 2024.

\bibitem{toolformer}
T.~Schick, J.~Dwivedi{-}Yu, R.~Dess{\`{\i}}, R.~Raileanu, M.~Lomeli, E.~Hambro, L.~Zettlemoyer, N.~Cancedda, and T.~Scialom, ``Toolformer: Language models can teach themselves to use tools,'' in \emph{NeurIPS}, 2023.

\bibitem{rag-survey}
Y.~Gao, Y.~Xiong, X.~Gao, K.~Jia, J.~Pan, Y.~Bi, Y.~Dai, J.~Sun, Q.~Guo, M.~Wang, and H.~Wang, ``Retrieval-augmented generation for large language models: {A} survey,'' \emph{CoRR}, vol. abs/2312.10997, 2023.

\bibitem{lpkg}
J.~Wang, M.~Chen, B.~Hu, D.~Yang, Z.~Liu, Y.~Shen, P.~Wei, Z.~Zhang, J.~Gu, J.~Zhou, J.~Z. Pan, W.~Zhang, and H.~Chen, ``Learning to plan for retrieval-augmented large language models from knowledge graphs,'' in \emph{{EMNLP} (Findings)}.\hskip 1em plus 0.5em minus 0.4em\relax Association for Computational Linguistics, 2024, pp. 7813--7835.

\bibitem{deepseek-r1}
DeepSeek{-}AI, D.~Guo, D.~Yang, H.~Zhang, J.~Song, R.~Zhang, R.~Xu, Q.~Zhu, S.~Ma, P.~Wang, X.~Bi, X.~Zhang, X.~Yu, Y.~Wu, Z.~F. Wu, Z.~Gou, Z.~Shao, Z.~Li, Z.~Gao, A.~Liu, B.~Xue, B.~Wang, B.~Wu, B.~Feng, C.~Lu, C.~Zhao, C.~Deng, C.~Zhang, C.~Ruan, D.~Dai, D.~Chen, D.~Ji, E.~Li, F.~Lin, F.~Dai, F.~Luo, G.~Hao, G.~Chen, G.~Li, H.~Zhang, H.~Bao, H.~Xu, H.~Wang, H.~Ding, H.~Xin, H.~Gao, H.~Qu, H.~Li, J.~Guo, J.~Li, J.~Wang, J.~Chen, J.~Yuan, J.~Qiu, J.~Li, J.~L. Cai, J.~Ni, J.~Liang, J.~Chen, K.~Dong, K.~Hu, K.~Gao, K.~Guan, K.~Huang, K.~Yu, L.~Wang, L.~Zhang, L.~Zhao, L.~Wang, L.~Zhang, L.~Xu, L.~Xia, M.~Zhang, M.~Zhang, M.~Tang, M.~Li, M.~Wang, M.~Li, N.~Tian, P.~Huang, P.~Zhang, Q.~Wang, Q.~Chen, Q.~Du, R.~Ge, R.~Zhang, R.~Pan, R.~Wang, R.~J. Chen, R.~L. Jin, R.~Chen, S.~Lu, S.~Zhou, S.~Chen, S.~Ye, S.~Wang, S.~Yu, S.~Zhou, S.~Pan, and S.~S. Li, ``Deepseek-r1: Incentivizing reasoning capability in llms via reinforcement learning,'' \emph{CoRR}, vol. abs/2501.12948, 2025.

\bibitem{baichuan-tech-report}
M.~Lin, F.~Yang, Y.~Shen, H.~Sun, T.~Li, T.~Zhang, C.~Zhu, T.~Zhang, M.~Zheng, X.~Li, Y.~Zhou, M.~Chen, Y.~Qin, Y.~Li, H.~Liang, F.~Li, Y.~Li, M.~Wang, G.~Dong, K.~Fang, J.~Xu, B.~Cui, W.~Zhang, Z.~Zhou, and W.~Chen, ``Baichuan alignment technical report,'' \emph{CoRR}, vol. abs/2410.14940, 2024.

\bibitem{qwen25-tech-report}
A.~Yang, B.~Yang, B.~Zhang, B.~Hui, B.~Zheng, B.~Yu, C.~Li, D.~Liu, F.~Huang, H.~Wei, H.~Lin, J.~Yang, J.~Tu, J.~Zhang, J.~Yang, J.~Yang, J.~Zhou, J.~Lin, K.~Dang, K.~Lu, K.~Bao, K.~Yang, L.~Yu, M.~Li, M.~Xue, P.~Zhang, Q.~Zhu, R.~Men, R.~Lin, T.~Li, T.~Xia, X.~Ren, X.~Ren, Y.~Fan, Y.~Su, Y.~Zhang, Y.~Wan, Y.~Liu, Z.~Cui, Z.~Zhang, and Z.~Qiu, ``Qwen2.5 technical report,'' \emph{CoRR}, vol. abs/2412.15115, 2024.

\end{thebibliography}

\end{document}